\documentclass[letterpaper, 10 pt, conference]{ieeeconf}  

\IEEEoverridecommandlockouts                              
\usepackage{graphics} 
\usepackage{epsfig} 
\usepackage{mathptmx} 
\usepackage{booktabs}
\usepackage{times} 
\usepackage{amsmath} 
\usepackage{amssymb}  
\usepackage{multirow}
\usepackage{gensymb}
\usepackage{color}
\usepackage{nicefrac}
\newcommand{\tabincell}[2]{\begin{tabular}{@{}#1@{}}#2\end{tabular}} 
\title{\LARGE \bf
ManhattanSLAM: Robust Planar Tracking and Mapping Leveraging Mixture of Manhattan Frames
}

\author{Raza Yunus$^{1}$, Yanyan Li$^{1}$ and Federico Tombari$^{1,2}$
\thanks{$^1$:Technical University of Munich, Germany; {\tt\small \{raza.yunus, yanyan.li, federico.tombari\}@tum.de} $^2$:Google Inc.}
\thanks{We gratefully acknowledge Stefano Gasperini for the helpful discussion. Yanyan Li is the corresponding author.}
}

\begin{document}

\maketitle
\thispagestyle{empty}
\pagestyle{empty}

\begin{abstract}
In this paper, a robust RGB-D SLAM system is proposed to utilize the structural information in indoor scenes, allowing for accurate tracking and efficient dense mapping on a CPU.
Prior works have used the Manhattan World (MW) assumption to estimate low-drift camera pose, in turn limiting the applications of such systems.
This paper, in contrast, proposes a novel approach delivering robust tracking in MW and non-MW environments. We check orthogonal relations between planes to directly detect Manhattan Frames, modeling the scene as a Mixture of Manhattan Frames. For MW scenes, we decouple pose estimation and provide a novel drift-free rotation estimation based on Manhattan Frame observations. For translation estimation in MW scenes and full camera pose estimation in non-MW scenes, we make use of point, line and plane features for robust tracking in challenging scenes.
Additionally, by exploiting plane features detected in each frame, we also propose an efficient surfel-based dense mapping strategy, which divides each image into planar and non-planar regions. Planar surfels are initialized directly from sparse planes in our map while non-planar surfels are built by extracting superpixels.
We evaluate our method on public benchmarks for pose estimation, drift and reconstruction accuracy, achieving superior performance compared to other state-of-the-art methods. We will open-source our code in the future.

\end{abstract}

\section{Introduction}
Environment-agnostic tracking and mapping based on an RGB-D camera play a central role in robotic and mixed/augmented reality applications. Such systems enable various interactive tasks, relying on accurate pose estimates and dense reconstruction.

Among traditional methods, feature-based approaches are more robust to illumination changes, compared to direct methods. Pure point-based methods~\cite{klein2007parallel, mur2015orb} lead to unstable performance in low-textured scenes. Robustness can be improved by adding other geometric features, like lines and planes, to the system~\cite{gomez2019pl,zureiki2008slam,zhang2019point}.  

Without the use of global bundle adjustment and loop closure~\cite{mur2015orb,mur2017orb}, small errors in pose estimation accumulate over time, causing drift in the camera pose trajectory. The former is computationally expensive, especially with large maps, and the latter works only if the agent revisits a location. Another approach for drift reduction is the use of the Manhattan/Atlanta World assumption to estimate rotation, given the fact that drift is mostly driven by inaccurate rotation estimations~\cite{straub2015real,zhou2016divide}. This technique has been employed by~\cite{kim2017visual,kim2018linear} and our previous works~\cite{Li2020SSLAM,li2020rgbd}, which exploit parallel and perpendicular relationships between geometric features in the scene.
These methods model the environment as a single global Manhattan World (MW) and make the assumption for every frame, which is very limiting.

\begin{figure}[t]
  \centering
  \includegraphics[scale=0.172]{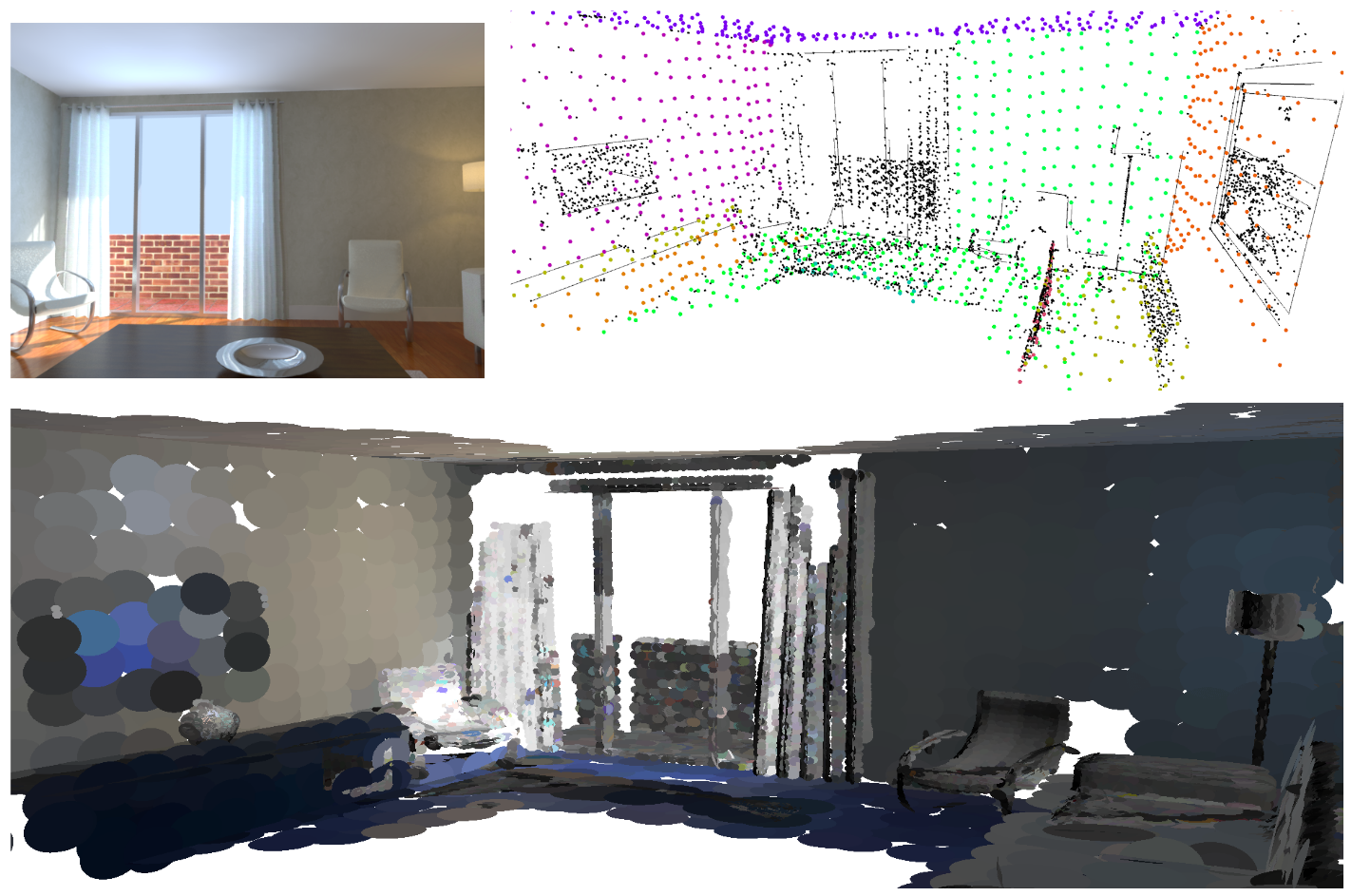}
  \caption[Sparse and dense reconstruction]{Indoor reconstruction from the proposed ManhattanSLAM framework. \textbf{Top Left:} Sample indoor scene. \textbf{Top Right:} Sparse reconstruction of the indoor environment. \textbf{Bottom:} Dense surfel-based reconstruction of the same environment.}\label{fig:reconstruction}
\end{figure}

In this paper, we alleviate the stringent requirement of the MW assumption by proposing a framework which can robustly utilize the MW structure, while also working in non-MW scenes, using point-line-plane tracking. We provide a method to detect Manhattan Frames (MF) directly from planes, allowing us to model the scene as a Mixture of Manhattan Frames (MMF)~\cite{straub2014mmf}, which is more applicable to real-world scenes, and estimate drift-free rotation by tracking MF observations across frames. Moreover, if no MFs are detected, our method switches to feature tracking, thus making it more robust than existing MW-based methods, as shown by our evaluation.

Additionally, to provide a dense map for robots, we propose an efficient surfel-based dense mapping strategy based on~\cite{wang2019real}. Different from \cite{Whelan2016ElasticFusionRD} and \cite{wang2019real}, where surfels are created for every pixel or superpixel, our method divides each keyframe into planar and non-planar regions. Surfels are initialized either from superpixels for non-planar regions or directly from our sparse map plane points for planar regions.
Therefore, compared to prior methods, the proposed strategy provides a more memory efficient dense reconstruction. The main contributions of this paper are summarized as:

i) A robust and general SLAM framework for indoor scenes, which takes the best of both worlds (MW assumption and feature tracking) by relying, when possible, on the MW structure for drift-free rotation estimation but able to seamlessly switch to feature tracking when MW does not hold.

ii) A novel drift-free rotation estimation method that tracks MF observations with the help of a Manhattan Map, generated by a suitable MF detection approach.


iii) An efficient dense surfel-based mapping strategy, which represents non-planar and planar regions by using superpixels and sparse plane points, respectively.

\section{Related Work}

ORB-SLAM~\cite{mur2015orb} is a popular point-based monocular SLAM system, which extends the multi-threaded and keyframe-based architecture of PTAM~\cite{klein2007parallel}. It uses ORB features, builds a co-visibility graph and performs loop closing and relocalization tasks. ORB-SLAM2~\cite{mur2017orb} further extends it to stereo and RGB-D sensors, while in ORB-SLAM3~\cite{campos2020orb}, inertial data, a multi-map module and support for an abstract camera model are integrated into the system.
To improve the robustness of point-based methods, lines and planes are extracted from the environment to deal with low/non-textured scenes.
\cite{zhang2011building} and \cite{zhou2015structslam} are extended from EKF-SLAM, building 3D line-based maps. \cite{zhang2015building} constructs a 3D line-based SLAM system using Pl\"ucker line coordinates for initialization and projection of 3D lines, and a 4 DoF orthonormal representation for optimization. Moreover, two recent homonymous techniques were proposed, PL-SLAM~\cite{gomez2019pl, pumarola2017pl}, which merge lines into a point-based system.
\cite{taguchi2013point} provides a RANSAC-based registration method for localization with hand-held 3D sensors, registering using points, planes, or a combination of them. In CPA-SLAM~\cite{ma2016cpa}, direct and dense DVO-SLAM~\cite{kerl2013dense} is extended to incorporate global plane landmarks for tracking the pose, in an Expectation-Maximization framework. \cite{kaess2015simultaneous} models infinite planes using the homogeneous parametrization and provides a minimum representation of planes for optimization, i.e., its azimuth angle, elevation angle and distance from the origin. Inspired by the MW assumption, SP-SLAM~\cite{zhang2019point} adds constraints between parallel and perpendicular planes in the scene.

Based on the MW assumption, \cite{zhou2016divide} proposes a mean-shift algorithm to track the rotation of MF across scenes, while using 1-D density alignments for translation estimation. OPVO~\cite{kim2017visual} improves the translation estimation by using the KLT tracker. Both methods require two planes to be visible in the frame at all times. LPVO~\cite{kim2018low} eases this requirement by incorporating lines into the system. Structural lines are aligned with the axes of the dominant MF and can be integrated into the mean shift algorithm, improving robustness. Hence for LPVO, only a single plane is required to be visible in the scene, given the presence of lines. Drift can still occur in translation estimation as it relies on frame-to-frame tracking. To tackle this, L-SLAM~\cite{kim2018linear} adds orthogonal plane detection and tracking on top of the LPVO architecture in a filtering framework. \cite{joo2020linear} extends the mean-shift algorithm for the more general scenario of Atlanta World, which can represent a wider range of scenes. \cite{Li2020SSLAM} allows the use of mean-shift algorithm for monocular scenes, by estimating surface normals from an RGB image using a convolutional neural network. \cite{li2020rgbd} further improves translation estimation by tracking plane features, in addition to points and lines, and adding parallel and perpendicular constraints between them.

KinectFusion~\cite{newcombe2011kinectfusion} provides an algorithm for real-time dense mapping and tracking of surfaces on a GPU using ICP alignment and a volumetric TSDF model. ElasticFusion~\cite{Whelan2016ElasticFusionRD} is another GPU-based approach that provides surfel-based maps of the environment. \cite{salas2014dense} builds a dense map using surfels, grouping them in large areas with little or no curvature to form planar regions that provide a semantic value for different applications. BundleFusion~\cite{dai2017bundlefusion} builds a globally consistent 3D reconstruction using a sparse-to-dense alignment strategy. Recently, real-time GPU-based mesh reconstruction techniques are proposed in \cite{schreiberhuber2019scalablefusion} and \cite{schops2019surfelmeshing}. \cite{wang2019real} proposes superpixel-based surfels to decrease the number of surfels in the map, which enables them to run their implementation on a CPU.

\section{Method}

\begin{figure*}[t]
  \centering
  \center{\includegraphics[scale=0.29]{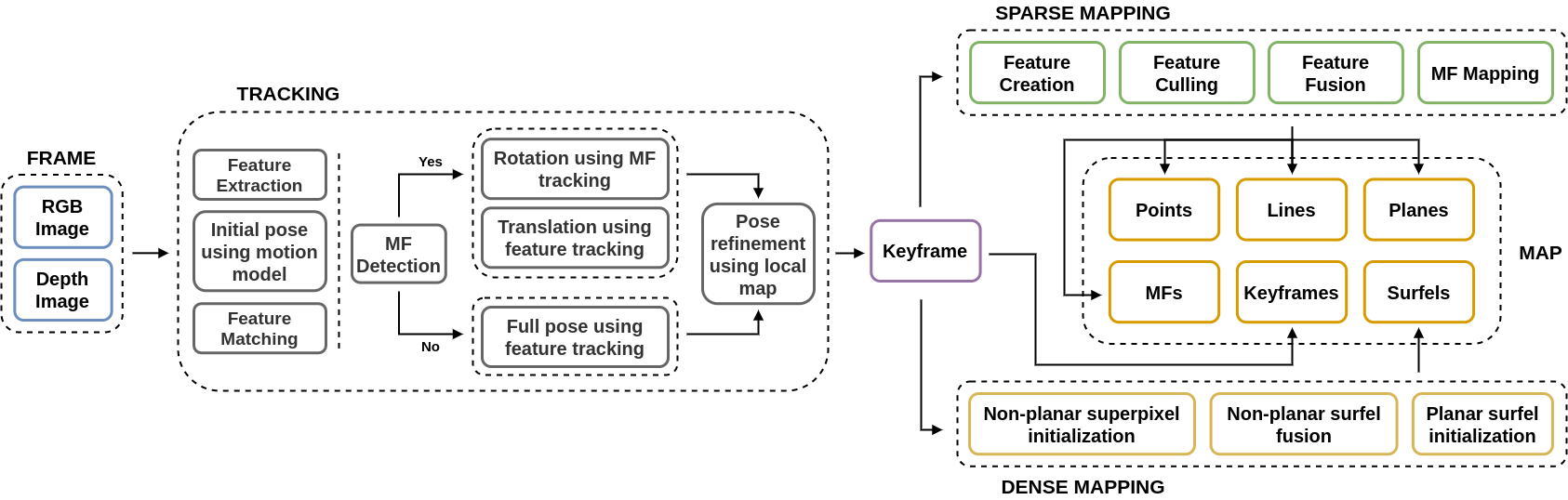}}
  \caption[System Overview]{Overview of the system showing three main tasks, i.e., tracking, sparse mapping and dense mapping, and components of the map.}\label{fig:pipeline}
\end{figure*}

Our system tackles three main tasks: tracking, sparse mapping and dense mapping, as shown in Figure \ref{fig:pipeline}. In this section, we explain the essential components of our system, including our novel approach for MF detection, drift-free rotation estimation and dense mapping.

\subsection{Tracking Framework}

For each RGB-D frame, points and lines are extracted from the RGB image while planes are extracted from the depth map. Similar to~\cite{mur2017orb}, we make use of the constant velocity motion model to get an initial pose estimate, that is further optimized by feature correspondences and structural regularities. For points and lines, a guided search from the last frame is used to match features, and planes are matched directly in the global map. Then, we detect MFs to determine whether the current scene is an MW scene or a non-MW scene, using the respective strategies for pose estimation, as described in Section~\ref{section:pose-estimation}. As an additional step in both cases, we track features in the local map of the current frame to further refine pose estimates. A new keyframe is created if the current frame observes less than 90\% of the points observed in the previous frame.

\subsection{Feature Detection and Matching}
\label{section:feature-detection}
Since points are difficult to extract in low-textured scenes, the proposed system also exploits the structural information of the environment, through lines and planes.

\subsubsection{Points}
For points, we use ORB features, which are based on the FAST keypoint detector~\cite{rosten2006machine} and BRIEF descriptor~\cite{calonder2010brief}. A 3D point is represented as $P=(X,Y,Z)$, while its 2D observation is represented as $p_{obs}=(u,v)$. Matches are determined by projecting 3D points on the image and finding the closest observation using Hamming distance between the respective descriptors.
\subsubsection{Lines}
To detect and describe line segments in the image, we use the robust LSD detector~\cite{von2008lsd} and the LBD descriptor~\cite{zhang2013efficient}. We represent 3D lines and their 2D observations with their endpoints $(P^l_{start}, P^l_{end})$ and $(p^l_{start}, p^l_{end})$ respectively while also obtaining normalized line function for the observed 2D line as $l_{obs} = \nicefrac{p^l_{start} \times p^l_{end}}{\|p^l_{start}\|\|p^l_{end}\|} = (a,b,c)$.
To determine a match between a 3D line and a 2D observation, both endpoints of the 3D line are individually projected and matched using the LBD descriptor.
\subsubsection{Planes}
\label{section:plane-detection}
Planes are extracted from the downsampled 3D point cloud using the AHC method~\cite{feng2014fast}, which provides the plane coefficients $(n, d)$ and supporting points in the point cloud for each plane instance. $n=(n_x,n_y,n_z)$ is the unit plane normal and $d$ is the distance of the plane from origin. We further downsample the point cloud of each plane using voxel-grid filtering, with a voxel size of $0.2 m$. Moreover, we discard potentially unstable plane observations, where the maximum point-plane distance between the plane and its points is larger than $0.04 m$.
For pose optimization, we use the minimal representation of planes: $q(\pi)=(\phi=arctan(\frac{n_y}{n_x}), \psi=arcsin(n_z), d)$,
where $\phi$ and $\psi$ are the azimuth and elevation angles of the plane normal. Plane matches are determined by comparing the angle between normals and the point-plane distance of planes.

\subsection{Manhattan Frame Detection and Mapping}
\label{section:mf-detection}

In contrast to using the popular mean-shift clustering algorithm~\cite{kim2018linear,Li2020SSLAM} for MF detection, which uses per-pixel normal estimation, we exploit the plane normals already extracted from the scene. An MF $M_k$ can be represented by three mutually perpendicular plane normals $(n^k_1, n^k_2, n^k_3)$. To detect an MF $M_k$ in the current frame $F_i$, we check the angle between the plane normals $n_z$, where $n_z \in \{n_0 ... n_r\}$ is the normal of a detected plane and $r$ is the total number of detected planes in $F_i$. An MF is detected whenever any three planes are mutually perpendicular. We can represent the observation of $M_k$ in camera coordinates $C_i$ of $F_i$ with a rotation matrix
\begin{equation}
R_{c_im_k} = \begin{bmatrix} n^k_1 & n^k_2 & n^k_3 \end{bmatrix}.
\end{equation}
 If only two perpendicular normals $n^k_1$ and $n^k_2$ are found, the third one $n^k_3$ can be recovered by taking the cross product between $n^k_1$ and $n^k_2$, thus the MF can be recovered from two planes as well.

Since sensor noise can lead to inconsistencies, where the columns of the matrix $R_{c_im_k}$ are not orthogonal, we use SVD to approximate $R_{c_im_k}$ with the closest rotation matrix $\hat{R}_{c_im_k}$:
\begin{equation}
SVD(R_{c_im_k}) \; = \; UDV^T,
\end{equation}
\begin{equation}
\hat{R}_{c_im_k} \; = \; UV^T.
\end{equation}

Furthermore, we also build a Manhattan map $G$ to collect MFs encountered in the scenes, where $G$ stores both full and partial MF observations along with the corresponding frames in which they are observed:
\begin{equation}
G = \{ M_k \mapsto F_i \}.
\end{equation}
Building this map allows us to estimate drift-free rotation when we encounter MF $M_k$ in any successive frame $F_j$.

To find a match between two observations of the same MF in camera frames $F_i$ and $F_j$, we check for matches of their constituent planes to the map planes. Each map plane has a unique ID in the map. 
If planes of both observations are matched to the same map planes, determined by comparing IDs, then the observations belong to the same MF.

\subsection{Pose Estimation}
\label{section:pose-estimation}

The camera pose $\xi_{cw}$ consists of a rotation matrix $R_{cw} \in SO(3)$ and a translation vector $t_{cw} \in \mathbb{R}^3$, from world coordinates $W$ to camera coordinates $C$. 
If MW assumption is not followed, determined by the failure to detect any MF, we estimate the full 6D pose by tracking features. In case of an MW scene, rotation and translation are decoupled and estimated separately.

\subsubsection{\textbf{For non-MW scenes}}
\label{section:non-mf-pose-estimation}
In non-MW scenes, points, lines and planes can be tracked to estimate a 6D camera pose. We define reprojection errors $e_p$, $e_l$ and $e_\pi$ between observed features and their corresponding matched 3D features in the map as
\begin{equation}
    \left\{
    \begin{array}{l}
        e_p \; = \; p_{obs} - \Pi (R_{cw} P_w + t_{cw}) \\
        e_l \; = \; {l_{obs}}^T \Pi (R_{cw} P^l_x + t_{cw}) \\
        e_{\pi} \; = \; q(\pi_c) - q (T_{cw}^{-T} \pi_w)
    \end{array},
    \right.
\end{equation}
where $\Pi$ is the projection function using the intrinsic camera matrix and $P^l_x$ is an endpoint of the 3D line, with $x \in \{start, end\}$. We also find parallel and perpendicular plane matches for each observed plane, which are added as structural constraints $e_{\pi_{\parallel}}$ and $e_{\pi_{\perp}}$ to the overall energy function as
\begin{equation}
    \left\{
    \begin{array}{l}
        e_{\pi_{\parallel}} \; = \; q_n(n_c) - q_n(R_{cw} n_w) \\
        e_{\pi_{\perp}} \; = \; q_n(R_{\perp} n_c) - q_n(R_{cw} n_w)
    \end{array},
    \right.
\end{equation}
where $n_c$ and $n_w$ are the normals of the observed plane and matched plane landmark, $R_{\perp}$ is a 90$\degree$ rotation matrix and $q_n(\pi)=(\phi, \psi)$.

Assuming a Gaussian noise model and combining all errors, the final energy function is written as $e = \sum \rho_y \left({e_y}^T \Lambda_y e_y\right)$, where $y \in \{p, l, \pi, \pi_{\parallel}, \pi_{\perp}\}$ and $\Lambda$ and $\rho$ denote the inverse covariance matrix and the robust Huber cost function, respectively. This energy function is optimized using the Levenberg-Marquardt algorithm to get the optimal pose estimate $\xi_{cw}^* = \operatorname*{argmin}_{\xi_{cw}} (e)$.

\subsubsection{\textbf{For MW scenes}}

In structured MW scenes, we decouple pose estimation and use our novel approach to estimate drift-free rotation, while feature tracking is used for translation estimation. For rotation estimation, all MFs in the scene can be detected using the method described in Section~\ref{section:mf-detection}. For each detected MF $M_k$ in frame $F_i$, represented by the corresponding rotation $R_{c_im_k}$, we search for $M_k$ in our Manhattan map $G$. If $M_k$ is found in G, we can obtain the corresponding frame $F_j$ from G, in which $M_k$ was first observed. $F_j$ serves as the reference frame, containing the MF observation $R_{c_jm_k}$ and pose estimate $\xi_{c_jw}$, which could have either been estimated by MF tracking or feature tracking.

Our goal here is to obtain the rotation $R_{c_iw}$ from world coordinates to current camera frame $F_i$. To achieve that, first, we use the observations of $M_k$ in $F_i$ and $F_j$ to calculate the relative rotation between them as
\begin{equation}
R_{c_jc_i} = R_{c_jm_k} {R_{c_im_k}}^T.
\end{equation}
Then, we take the rotation estimate $R_{c_jw}$ from the pose estimate $\xi_{c_jw}$ of $F_j$ and concatenate it with the relative rotation between $F_i$ and $F_j$ to get
\begin{equation}
R_{wc_i} = {R_{c_jw}}^T R_{c_jc_i}.
\end{equation}
Finally, we transpose $R_{wc_i}$ to get our desired rotation $R_{c_iw}$.

Such rotation estimation is only possible if $M_k$ has been already observed, i.e. $M_k$ is stored in $G$. If $F_i$ does not contain any previously observed MF, then we use the feature tracking method (Section~\ref{section:non-mf-pose-estimation}) to estimate the full pose. When an MF $M_k$ is observed for the first time, we store it in our Manhattan map $G$, along with its frame $F_i$. For any subsequent observations of $M_k$, $F_i$ can be used as a reference for rotation estimation. In case of multiple MF detections in $F_i$, we select the dominant MF, i.e. the one which contains the highest number of points in the point clouds of its constituent planes.

\begin{figure}[t]
  \centering
  \center{\includegraphics[scale=0.36]{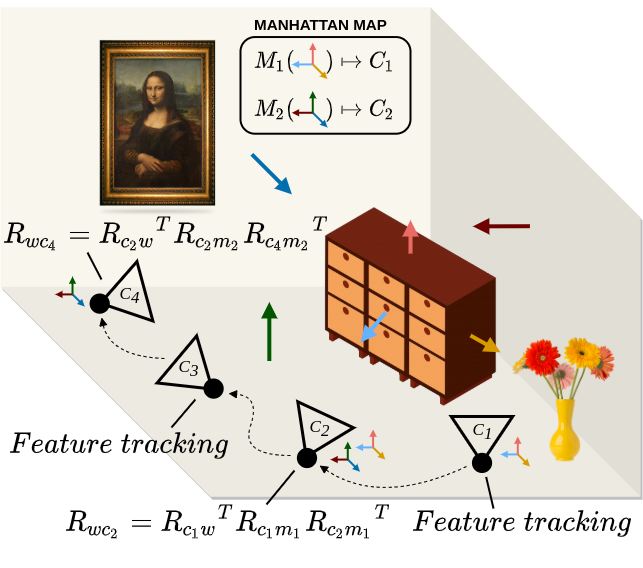}}
  \caption[MF Tracking]{A toy example showing two MFs and estimated pose for four camera frames. $C_1$ observes MF $M_1$, which is added to the Manhattan Map G. It uses feature tracking as there was no previous observation of $M_1$. $C_2$ observes both $M_1$ and $M_2$, so $M_2$ is added to G. Since $M_1$ is already present in G, a drift-free rotation is estimated for $C_2$ with $C_1$ as reference frame. $C_3$ does not observe any MF while $C_4$ observes $M_2$, which allows for a drift-free rotation using $C_2$ as reference frame.}\label{fig:mf-tracking}
\end{figure}

Now that we have the rotation $R_{cw}$, we want to find the corresponding translation $t_{cw}$ that will give us the full camera pose $\xi_{cw}$. For this, we use feature tracking as described in Section~\ref{section:non-mf-pose-estimation}. The translation can be obtained by solving $t_{cw}^* = \operatorname*{argmin}_{t_{cw}} \; (e_t)$, where $e_t = \sum \rho_z \left({e_z}^T \Lambda_z e_z\right)$ and $z \in \{p, l, \pi\}$. We fix rotation and only update the translation during the optimization process. Note that we do not use parallel and perpendicular constraints for planes here, since they only provide rotational information.
\subsection{Sparse mapping}
Our SLAM system maintains a sparse map of landmarks and keyframes. 
For our sparse map, we follow the keyframe-based approach of \cite{mur2015orb}, where a new frame is only added when it observes a significant number of new landmarks. New landmarks, i.e. points, lines and planes, are initialized and added to the map from keyframes using the depth map provided by the RGB-D image. If a new plane is matched to a previous map plane, we only update the point cloud of the map plane, otherwise, the new plane is added to the map. Following \cite{mur2015orb}, we maintain a co-visibility graph among keyframes to determine the local map of the current frame and remove unreliable landmarks and redundant keyframes using respective culling strategies.

\subsection{Dense surfel mapping}
To improve the reconstruction efficiency, we provide a novel dense mapping strategy based on \cite{wang2019real}. 
Instead of building a surfel for every pixel like ElasticFusion, \cite{wang2019real} divides each image into superpixels and builds surfels based on the extracted superpixels. This approach reduces the number of surfels, allowing it to run on a CPU.

In our method, we further improve the efficiency of \cite{wang2019real} by utilizing extracted planes in the scene. For planar regions, we build surfels by reusing planes from our sparse map, making our method more memory-efficient.
We update the method provided by \cite{wang2019real} as follows:
\begin{itemize}
    \item Our plane detection method provides a mask for planar regions in the frame. We use this mask to generate superpixels for non-planar regions, using the modified SLIC~\cite{achanta2012slic} method of \cite{wang2019real}.
    \item Surfels are generated and fused for non-planar regions using the method of \cite{wang2019real}.
    \item For planar regions, we use the points from our sparse planes as surfel positions. Each surfel is assigned the normal of the corresponding plane. To determine radius of the surfel, we utilize the size of the voxel used to downsample our plane during voxel grid filtering. We take the length of the cross sectional diagonal of the voxel, divide it by two and set that as the radius.
\end{itemize}

\begin{figure}[t]
  \centering
  \center{\includegraphics[scale=0.145]{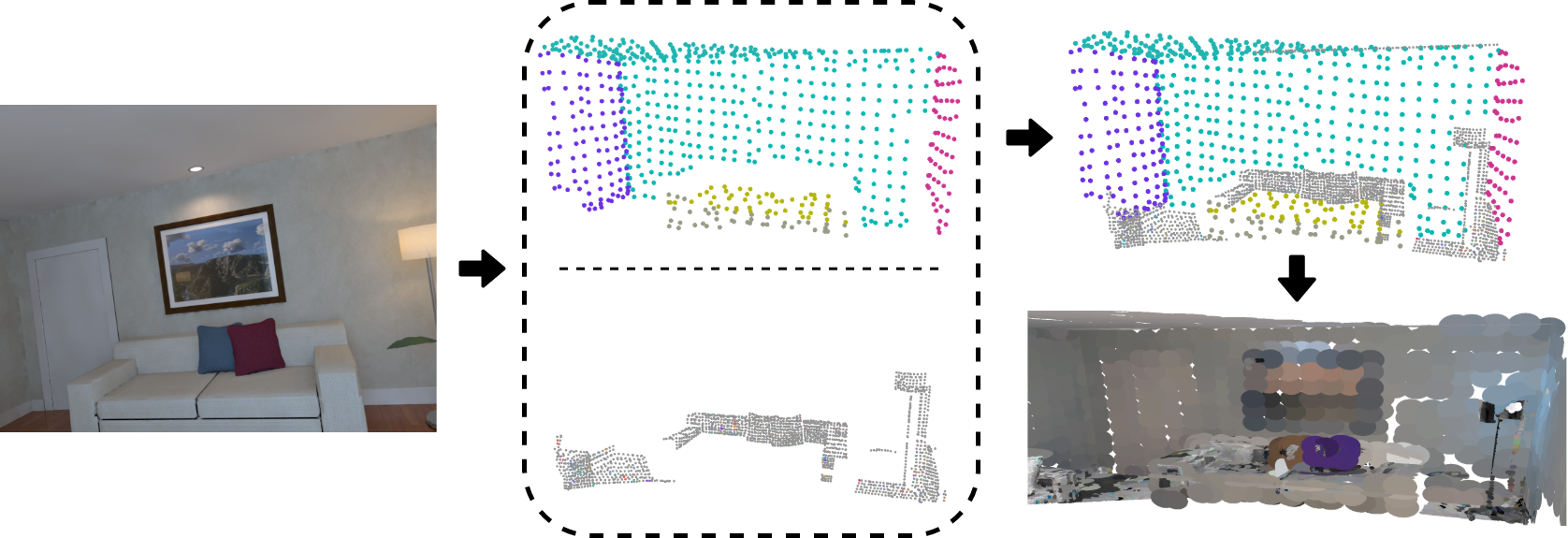}}
  \caption[Dense Mapping]{\textbf{Left:} Sample scene. \textbf{Middle:} Points in sparse map planes and non-planar surfel points of the same scene. \textbf{Top Right:} Combined plane points and surfel points. \textbf{Bottom Right:} Rendered surfels.}\label{fig:dense-mapping}
\end{figure}


\section{Evaluation}

In this section, we evaluate multiple aspects of our system on publicly available datasets and 
compare it with feature-based methods ORB-SLAM2 and SP-SLAM, MW-based method L-SLAM and our previous MW-based works S-SLAM and RGBD-SLAM. All experiments are performed on an Intel Core i5-8250U CPU @ 1.60GHz $\times$ 8 with 19.5 GB RAM. We do not use any GPU for our experiments. Our method runs at around 15 Hz, taking 67 ms for tracking and 40 ms for superpixel extraction and surfel fusion (on a separate thread), on average. Additionally, we disable the bundle adjustment and loop closure modules of ORB-SLAM2 and SP-SLAM for a fair comparison.

\subsection{Pose Estimation}

\subsubsection{ICL-NUIM}
The ICL-NUIM~\cite{handa2014benchmark} dataset provides camera sequences containing scenes for two synthetically-generated indoor environments: a living room and an office room. These environments contain large areas of low-texture surfaces like walls, ceiling, and floor. 
Table \ref{table:icl} shows the performance of our method based on translation ATE RMSE, compared to other feature- and MW-based SLAM systems. We also show the number of frames where MF tracking was used. Since ICL-NUIM is rendered based on a rigid Manhattan World model, MW-based methods work well, specially L-SLAM in of-kt0 and of-kt3 sequences and RGBD-SLAM~\cite{li2020rgbd} in lr-kt0 and of-kt2.
However, MW-based methods are sensitive to the structure of environment as they need two perpendicular elements for every scene. In living room environments, especially in lr-kt3, some viewpoints are too close to the wall and contain only one plane, which leads to errors for MW-based approaches. Our method, however, is more robust as it switches to feature tracking in these cases, as well as in scenes where the detected planes are noisy.
Feature-based methods ORB-SLAM and SP-SLAM also work well as both environments contain abundant texture. Nevertheless, our approach outperforms prior methods, by taking advantage of both structure and texture in the scene.


\begin{table*}[h]
\centering
\caption[ATE comparison]{Comparison of ATE RMSE (m) for ICL-NUIM and TUM RGB-D sequences. $\times$ represents tracking failure. - means result is not available.}
 \begin{tabular}{c|c|cccccc|cc}
\toprule
 & & \multicolumn{6}{c|}{Methods} & \multicolumn{2}{c}{Frames} \\ \hline
 Dataset & Sequence & Ours & S-SLAM~\cite{Li2020SSLAM} & RGBD-SLAM~\cite{li2020rgbd} & ORB-SLAM2~\cite{mur2017orb} & SP-SLAM~\cite{zhang2019point} & L-SLAM~\cite{kim2018linear} & Total & MF\\
 \hline
 \multirow{9}{*}{\tabincell{c}{ICL\\NUIM}} & lr-kt0 & 0.007 & - & \textbf{0.006} & 0.014 & 0.019 & 0.015 & 1510 & 1203 \\ 
 & lr-kt1 & \textbf{0.011} & 0.016 & 0.015 & \textbf{0.011} & 0.015 & 0.027 & 967 & 678 \\ 
 & lr-kt2 & \textbf{0.015} & 0.045 & 0.020 & 0.021 & 0.017 & 0.053 & 882 & 771 \\ 
 & lr-kt3 & \textbf{0.011} & 0.046 & 0.012 & 0.018 & 0.022 & 0.143 & 1242 & 1030 \\
 & of-kt0 & 0.025 & - & 0.041 & 0.049 & 0.031 & \textbf{0.020} & 1510 & 1505 \\ 
 & of-kt1 & \textbf{0.013} & $\times$ & 0.020 & 0.029 & 0.018 & 0.015 & 967 & 945 \\ 
 & of-kt2 & 0.015 & 0.031 & \textbf{0.011} & 0.030 & 0.027 & 0.026 & 882 & 786 \\
 & of-kt3 & 0.013 & 0.065 & 0.014 & 0.012 & 0.012 & \textbf{0.011} & 1242 & 1212 \\ \hline
 & Average & \textbf{0.014} & 0.040 & 0.017 & 0.023 & 0.020 & 0.039 & & \\ \hline
 \multirow{10}{*}{\tabincell{c}{TUM \\RGB-D}} & fr1/xyz & \textbf{0.010} & $\times$ & $\times$ & \textbf{0.010} & \textbf{0.010} & - & 798 & 1 \\
 & fr1/desk & 0.027 & $\times$ & $\times$ & \textbf{0.022} & 0.026 & - & 613 & 1 \\
 & fr2/xyz & \textbf{0.008} & $\times$ & $\times$ & 0.009 & 0.009 & - & 3669 & 0 \\
 & fr2/desk & 0.037 & $\times$ & $\times$ & 0.040 & \textbf{0.025} & - & 2965 & 26 \\
 & fr3/s-nt-far & 0.040 & 0.281 & \textbf{0.022} & $\times$ & 0.031 & 0.141 & 793 & 688 \\
 & fr3/s-nt-near & \textbf{0.023} & 0.065 & 0.025 & $\times$ & 0.024 & 0.066 & 1053 & 796 \\
 & fr3/s-t-near & 0.012 & 0.014 & - & 0.011 & \textbf{0.010} & 0.156 & 1056 & 564 \\
 & fr3/s-t-far & 0.022 & 0.014 & - & \textbf{0.011} & 0.016 & 0.212 & 906 & 576 \\
 & fr3/cabinet & \textbf{0.023} & - & 0.035 & $\times$ & $\times$ & 0.291 & 1111 & 985 \\
 & fr3/l-cabinet & 0.083 & - & \textbf{0.071} & $\times$ & \textbf{0.074} & 0.140 & 983 & 188 \\
 \bottomrule
\end{tabular}
\label{table:icl}
\end{table*}

\subsubsection{TUM RGB-D}
The TUM RGB-D benchmark~\cite{sturm2012benchmark} is another popular dataset for the evaluation of SLAM algorithms. It consists of several real-world camera sequences which contain a variety of scenes, like cluttered areas and scenes containing varying degrees of structure and texture. MW-based systems struggle in cluttered environments, while point-based systems perform poorly in scenes lacking texture, so such a variety of scenes is suitable for showcasing how our system can robustly adapt to both MW and non-MW scenes.

In the fr1 and fr2 sequences where scenes are cluttered and contain few or no MFs, MW-based methods S-SLAM, RGBD-SLAM and L-SLAM cannot track as they need an MF for every frame, as shown in Table \ref{table:icl}. Instead, the proposed method can robustly estimate pose in non-MW scenes, performing equivalently to feature-based ORB-SLAM and SP-SLAM. For the fr3 sequence, our decoupled MW-based estimation gives improved results for structured environments. Four of the six tested sequences contain no or limited texture, resulting in a failure of ORB-SLAM2. SP-SLAM uses plane features as well, so it provides good results on all sequences except for 'cabinet'. On the other hand, MW-based S-SLAM and L-SLAM exploit structural information, although the lack of texture affects their translation estimation. RGBD-SLAM uses planes and structural constraints for translation estimation as well, so it works particularly well for 's-nt-far' and 'l-cabinet' sequences. As depth data in the TUM RGB-D sequences is captured from real-world scenes, it is not as accurate as sequences from the ICL-NUIM dataset. Hence, MW-based methods suffer due to noisy surface normals, especially in the cabinet and large-cabinet sequences. This affects our method as well, so to circumvent this, our method switches to feature tracking for frames with noisy planes.


\subsection{Drift}

To test the amount of accumulated drift and robustness over time, we evaluate our system on the TAMU RGB-D dataset~\cite{lu2015robust} which contains long indoor sequences. Although the dataset does not provide ground-truth poses, the camera trajectory is a loop, so we can calculate the accumulated drift by taking the Euclidean distance between the starting and ending point of our estimated trajectory.
\begin{table}[t]
    \centering
    \caption[TAMU RGB-D - Drift comparison]{Comparison of the accumulated drift (m) in TAMU RGB-D sequences. -MF means only feature tracking is used.}
    \begin{tabular}{l|ccc|cc}
    \toprule
    & & & & \multicolumn{2}{c}{Frames} \\ \hline
    Sequences & Ours & Ours/-MF & ORB-SLAM2~\cite{mur2017orb} & Total & MF\\ \hline
    Corridor-A & 0.53 & 0.77 & 3.13 & 2658 & 401\\
    Entry-Hall & 0.39 & 0.81 & 2.22 & 2260 & 282\\
    \bottomrule
    \end{tabular}
    \label{table:tamu}
\end{table}
\begin{figure}[t]
  \centering
  \center{\includegraphics[scale=0.35]{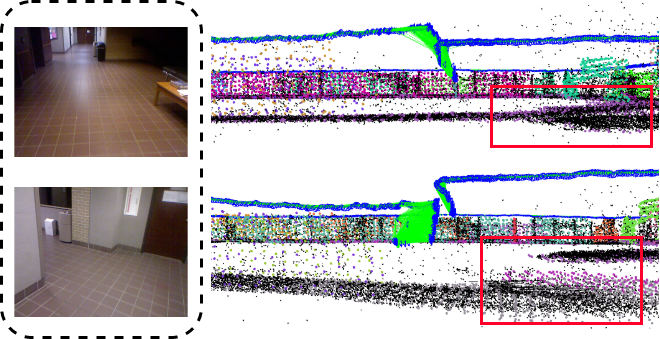}}
  \caption[Drift]{Drift for TAMU-RGBD Corridor-A sequence. The pose estimates and reconstruction of our method around the loop point are shown. The red boxes highlight the inconsistency in ground plane when drift-free rotation is not used. \textbf{Left:} Sample scenes from the sequence. \textbf{Top Right:} Result with drift-free rotation enabled. \textbf{Bottom Right:} Result with drift-free rotation disabled.}\label{fig:drift}
\end{figure}

Table \ref{table:tamu} shows the drift of our method, compared to ORB-SLAM2. Since TAMU RGB-D dataset has real-world scenes with noisy depth data, our method uses drift-free MF tracking only for frames with less noisy planes. We also evaluate the effect of our MF tracking method on the drift in pose estimates. Without MF tracking, our method still performs better than ORB-SLAM2, thanks to the addition of planes and structural constraints in the feature tracking module. With the addition of MF tracking proposed in our method, the drift of pose estimation is further reduced. It can be seen in Figure~\ref{fig:drift} that the reconstruction of floor aligns better at the loop point when MF tracking is enabled. These results indicate that drift could be further reduced with less noisy depth data, as it would result in more MFs being detected and used for drift-free rotation estimation.

\subsection{Reconstruction Accuracy}

\begin{table}[t]
\small
    \centering
    \caption[Reconstruction results]{Reconstruction error (cm) on the ICL-NUIM dataset.}
    \begin{tabular}{c|ccccc}
       \hline
        Sequence & E-Fus~\cite{Whelan2016ElasticFusionRD}  & InfiniTAM~\cite{prisacariu2017infinitam} & DSM~\cite{wang2019real}  & Ours \\ \hline
        lr-kt0   & 0.7 & 1.3  & 0.7 & \textbf{0.5} \\
        lr-kt1   & 0.7 & 1.1  & 0.9          & \textbf{0.6} \\
        lr-kt2   & 0.8  & \textbf{0.1} & 1.1          & 0.7 \\
        lr-kt3   & 2.8 &2.8  &1.0 & \textbf{0.7}\\ \hline 
    \end{tabular}
    \label{table:reconstruction}
\end{table}

Table~\ref{table:reconstruction} shows the reconstruction accuracy of our method evaluated on the living room sequences of the ICL-NUIM dataset. The evaluation is based on the point cloud generated by our surfels. ElasticFusion and InfiniTAM show good performance, with the latter getting an excellent result for lr-kt2. DSM~\cite{wang2019real}, based on ORB-SLAM, performs admirably as the living room sequences have plenty of texture. Our method, however, uses structure in the environment and performs best on three out of four sequences. ElasticFusion and InfiniTAM need a GPU while DSM and our method work only on a CPU.




\section{Conclusion}

In this paper, we propose ManhattanSLAM, a method that tracks camera pose robustly in general indoor scenes, with the added ability of exploiting the structural regularities in MW scenes to compute low-drift pose estimates, as shown in our experiments.
Furthermore, we exploit planar regions in the scene to provide an efficient surfel-based dense reconstruction of the environment. Future enhancements to the system can include adding a loop closure module, improving plane detection to further discard unstable observations and making the planar surfel radius flexible to more closely fit the actual plane boundary.

{\small
\bibliographystyle{ieee}
\bibliography{egbib}
}

\end{document}